\documentclass[letterpaper]{article} 
\usepackage{aaai25}  
\usepackage{times}  
\usepackage{helvet}  
\usepackage{courier}  
\usepackage{amsfonts}
\usepackage[hyphens]{url}  
\usepackage{graphicx} 
\usepackage{multirow}
\usepackage{natbib}  
\usepackage{caption} 
\usepackage{algorithm}
\usepackage{algorithmic}
\usepackage{booktabs}
\usepackage{amsmath}
\usepackage{newfloat}
\usepackage{listings}
\DeclareCaptionStyle{ruled}{labelfont=normalfont,labelsep=colon,strut=off} 
\floatstyle{ruled}
\newfloat{listing}{tb}{lst}{}
\floatname{listing}{Listing}
\title{ColonScopeX: Leveraging Explainable Expert Systems with Multimodal Data for Improved Early Diagnosis of Colorectal Cancer}
\author{
    Natalia Sikora\textsuperscript{\rm 1},  Robert L. Manschke, Alethea M. Tang\textsuperscript{\rm 2},  \\ Peter Dunstan\textsuperscript{\rm 1},  Dean A. Harris\textsuperscript{\rm 2},  Su Yang\textsuperscript{\rm 3}
}
\affiliations{
    \textsuperscript{\rm 1} Department of Physics, Swansea University, United Kingdom\\
    \textsuperscript{\rm 2}Swansea University Medical School, Swansea University, United Kingdom\\    
    \textsuperscript{\rm 3}Department of Computer Science, Swansea University, United Kingdom\\ nmrsikora@gmail.com, su.yang@swansea.ac.uk
}

\begin{document}
\maketitle

\begin{abstract}
Colorectal cancer (CRC) ranks as the second leading cause of cancer-related deaths and the third most prevalent malignant tumour worldwide. Early detection of CRC remains problematic due to its non-specific and often embarrassing symptoms, which patients frequently overlook or hesitate to report to clinicians. Crucially, the stage at which CRC is diagnosed significantly impacts survivability, with a survival rate of 80-95\% for Stage I and a stark decline to 10\% for Stage IV. Unfortunately, in the UK, only 14.4\% of cases are diagnosed at the earliest stage (Stage I).

In this study, we propose ColonScopeX, a machine learning framework utilizing explainable AI (XAI) methodologies to enhance the early detection of CRC and pre-cancerous lesions. Our approach employs a multimodal model that integrates signals from blood sample measurements, processed using the Savitzky-Golay algorithm for fingerprint smoothing, alongside comprehensive patient metadata, including medication history, comorbidities, age, weight, and BMI. By leveraging XAI techniques, we aim to render the model's decision-making process transparent and interpretable, thereby fostering greater trust and understanding in its predictions. The proposed framework could be utilised as a triage tool or a screening tool of the general population.  

This research highlights the potential of combining diverse patient data sources and explainable machine learning to tackle critical challenges in medical diagnostics.
\end{abstract}
\section{Introduction}
Colorectal cancer (CRC) remains the second most deadly and third most common malignant tumour worldwide \cite{Tortora2022}.

Detecting CRC is challenging due to its non-specific and hard-to-identify symptoms \cite{Lam2021}. Patients frequently ignore or avoid discussing these symptoms because of discomfort, further complicating diagnosis \cite{Cossu2018}. Current CRC tests are either expensive, ineffective, or have low compliance due to their inconvenience. Ideally, testing should be non-invasive, easy to interpret, and capable of early detection \cite{Chan2022}.

This project focuses on using explainable AI (XAI) approaches to justify the decisions of a machine learning (ML) model designed to identify patients with polyps or early-stage disease based on signals obtained from blood samples.

The introduction of artificial intelligence (AI) in the clinic brings us closer to a future of medicine, where  population-wide implementation of personalised medicine is common practice. To advance towards personalised medicine and precision oncology, clinicians will need to utilise better data analysis approaches, including integrated multimodal datasets \cite{Li2024}. Although clinical investigations tend to produce immense amounts of data, working with clinical data is associated with substantial barriers, mainly due to data sparsity and scarcity \cite{Cui2023}.

The application of deep learning (DL) models in clinical settings faces significant challenges, including inconsistent data collection practices, budget constraints, and limited modalities \cite{Dinsdale2022}. These issues are often mitigated using imputation, interpolation, and matrix completion techniques \cite{Li2024-sn}.

However, clinical data itself presents additional hurdles, such as incomplete datasets, small cohort sizes, and a lack of high-quality multimodal dataset annotations. In some countries, a bias is introduced when patients primarily come from higher socioeconomic backgrounds, leading to model overfitting and poor generalisation \cite{dElia2022}.

Moreover, any AI application in a clinical setting must be transparent and easily explainable to medical professionals. It should also undergo rigorous testing and provide uncertainty and confidence measures to support its predictions \cite{Alowais2023-qd}.

This paper introduces the first multimodal framework capable of detecting both CRC and pre-cancerous disease from non-invasive and cost-effective datasets:
\begin{itemize}
\item We benchmark multiple algorithms on the dataset with two modalities; with the best-performing framework, ColoScopeX, reporting performance metrics for early, late, and joint fusion.
\item The results are presented in an easily interpretable text format,  facilitating the transfer of knowledge between computational specialists and clinicians.
\item This approach enables cost-effective and rapid screening of the general population, potentially reducing CRC-related deaths and eliminating the need for expensive laboratory techniques like circulating tumour DNA (ctDNA) data, exosome analysis, or extensive mass spectrometry (MS) panels.
\item The proposed method  brings us closer to personalised medicine by accounting for factors such as poly-medication, comorbidities, smoking status, age, sex, and other clinical characteristics.
\end{itemize}

\section{Background/Related Work}

\subsubsection{Cancer detection from single modality.} Recent advances in cancer detection techniques often rely on sophisticated laboratory methods that are costly, time-consuming, and require highly skilled personnel, particularly in ctDNA-based approaches \citep{Vittone2024}. In contrast, the literature presents Raman spectroscopy as a cost-effective, rapid, and straightforward alternative \citep{Hanna2021}. Raman-based techniques include chip-based methods that necessitate exosome extraction \citep{Shin2023}, tissue analysis methods that require surgical resection or biopsy \citep{Hanna2021}, and surface-enhanced Raman spectroscopy (SERS), which depends on biocompatible SERS tags \citep{Auner2018}. Among these, one of the most promising techniques reported by Shin et al. (2023) achieved an average AUC value of 0.925, with a sensitivity of 87.4\% and specificity of 88.3\%. Despite these impressive results, single-modality cancer detection methods present several challenges, including the need for highly qualified laboratory staff, expensive consumables, time-consuming procedures with strict quality controls, low biomarker abundance, degradation of biomarkers, and the use of toxic chemicals \cite{Sebastian2022}. In this paper, we propose an application of standard Raman spectroscopy that eliminates the need for complex laboratory preparations.

\subsubsection{Multimodal cancer detection and Raman Spectroscopy.} Literature findings show an improvement in cancer detection when using multiple modalities \citep{Tan2022}. Raman spectroscopy-based techniques utilizing multimodal datasets tend to be complicated, and appropriate clinical introduction necessitates more cost-effective and simpler techniques. \citet{Novikov2024} proposed a multimodal fiber probe for simultaneous mid-infrared and Raman spectroscopy, while \citet{Wang2023} introduced a CNN based on the Raman spectra of serum and clinical features, including patient age and PSA levels.

\citet{Wang2023} used SERS, utilizing silver nanoparticles (AgNPs). The multimodal approach they employed allowed for the identification of specific amino acids and lipids in lipidomics, differentiating prostate cancer from benign prostatic hyperplasia, potentially reducing the risk of patient overtreatment. With SERS-only data, the CNN model achieved a classification accuracy of 85.14\% with an AUC of 0.87. When incorporating SERS data in a multimodal CNN, classification accuracy improved to 88.55\% and the AUC to 0.91, outperforming traditional biomarkers, thus proving the potential to enhance prostate cancer diagnostics.

Although the approach suggested by \citet{Wang2023} is promising, the study has several limitations. When attempting to link the Gleason score and serum metabolites, there is a risk of false negatives in the biopsy procedure, and the spectral pattern obtained in this manner may be subject to inherent variability. Patients taking any medications between blood tests were excluded and were required to fast before sample collection. While this project demonstrated the potential of SERS-based CNN models in improving prostate cancer diagnostics—and this technique could be adapted for identifying other cancers—we present an improvement via a simpler Raman spectroscopy approach that examines only blood spectra fingerprints. Additionally, the multimodal approach in our paper includes a list of medications taken by the patient, additional symptoms, age, sex, and comorbidities.

\subsubsection{Optimised Pre-Processing of Raman Spectra for CRC Detection.} 
Signal pre-processing of spectra is a critical step in the clinical application of Raman spectroscopy. The framework presented in this study demonstrates an improvement over standard Raman spectroscopy approaches by focusing on optimizing pre-processing techniques for CRC detection. We based our work on \citet{Woods2022}, who proposed pre-processing methods that improved sensitivity by 14.6\%, specificity by 6.9\%, positive predictive value by 3.4\%, and negative predictive value by 2.4\% compared to standard pre-processing methods. We expanded on the recommendations provided by \citet{Woods2022} and introduced an additional data-cleaning step in both the pre- and post-processing stages.

\subsubsection{Current Liquid Biopsy Recommendations.} The Liquid Biopsy Consortium Updates summarised the current state-of-the-art in early cancer detection. Most techniques focus on the isolation of disease-specific analytes such as circulating tumor cells (CTCs), ctDNA, and extracellular vesicles (EVs) \citep{Batool2023}.

This consortium highlighted challenges related to the lack of standardised protocols, inconsistent use of isolation kits, and inadequate sample handling methods. Furthermore, they pointed out that study population selection often relies on convenience sampling, leading to suboptimal control groups, and that pre-sampling factors such as circadian rhythm and metabolic conditions affect analyte quality \citep{Batool2023}. These issues could be addressed by utilising a simpler laboratory technique, such as the framework proposed in our work.

\subsection{Dataset}
Our dataset comprises two modalities: a) Raman spectroscopy fingerprint readings obtained from patients' serum samples, and b) the corresponding patient metadata, including any medications taken, comorbidities, smoking status, and demographic information such as age, sex, and BMI. We selected 1,035 samples for this study, as shown in Table~\ref{tab:diagnosis_distribution}. Patients' CRC or polyp diagnoses were confirmed via colonoscopy, and each control participant had a six-month follow-up to exclude any misdiagnosis.


\begin{table}[t]
    \centering
    \setlength{\tabcolsep}{1mm}
    \small
    \begin{tabular}{lcc}
        \toprule
        Diagnosis & Sex & Count \\
        \midrule
        Control (Diagnosis 1) & M & 249 \\
        Control (Diagnosis 1) & F & 222 \\
        Polyp (Diagnosis 2) & M & 182 \\
        Polyp (Diagnosis 2) & F & 120 \\
        Early Stage Cancer (Diagnosis 0) & M & 149 \\
        Early Stage Cancer (Diagnosis 0) & F & 113 \\
        \bottomrule
    \end{tabular}
    \caption{Distribution of Patients by Sex in the Unbalanced Model. Summary of samples that passed the post and pre-processing filtering criteria}
    \label{tab:diagnosis_distribution}
\end{table}

Each of the patient samples in this dataset was measured at least 6 times within different locations of the sample, providing us with over 6210 different spectral measurements.

\begin{table}[t]
    \centering
    \setlength{\tabcolsep}{1mm}
    \small
    \begin{tabular}{lcc}
        \toprule
        Diagnosis & Smoking status & Count\\
        \midrule
        Control (Diagnosis 1) & 0 & 140 \\
        Early Stage Cancer (Diagnosis 0) & 0 & 139 \\
        Polyp (Diagnosis 2) & 0 & 109 \\
        Polyp (Diagnosis 2) & 1 & 75 \\
        Early Stage Cancer (Diagnosis 0) & 1 & 65 \\
        Control (Diagnosis 1) & 1 & 60 \\
        Polyp (Diagnosis 2) & 2 & 38 \\
        Control (Diagnosis 1) & 2 & 25 \\
        Early Stage Cancer (Diagnosis 0) & 2 & 22 \\
        Polyp (Diagnosis 2) & 4 & 4 \\
        Control (Diagnosis 1) & 4 & 1 \\
        \bottomrule
    \end{tabular}
    \caption{Diagnoses and the smoking status in samples which passed the post and pre-processing used for balanced out model (113 patients in each sex and diagnosis subtype).}
    \label{tab:balan_diagnosis_distribution}
\end{table}

We utilised two separate approaches, working on an imbalanced model and a balanced out model (Table~\ref{tab:diagnosis_distribution} and ~\ref{tab:balan_diagnosis_distribution}). 

\section{Method}
Patients with CRC and polyps present different metabolic profiles \citep{DiCesare2023}. Due to the size restrictions of our dataset and the differences in features, each sample is processed through two models: one trained to identify polyps only and one trained to identify CRC only. Both models share the same architecture. The clinician receives a report specifying whether the sample was classified as diseased by either of these models.

\begin{figure*}[t]
\centering
\includegraphics[width=\textwidth]{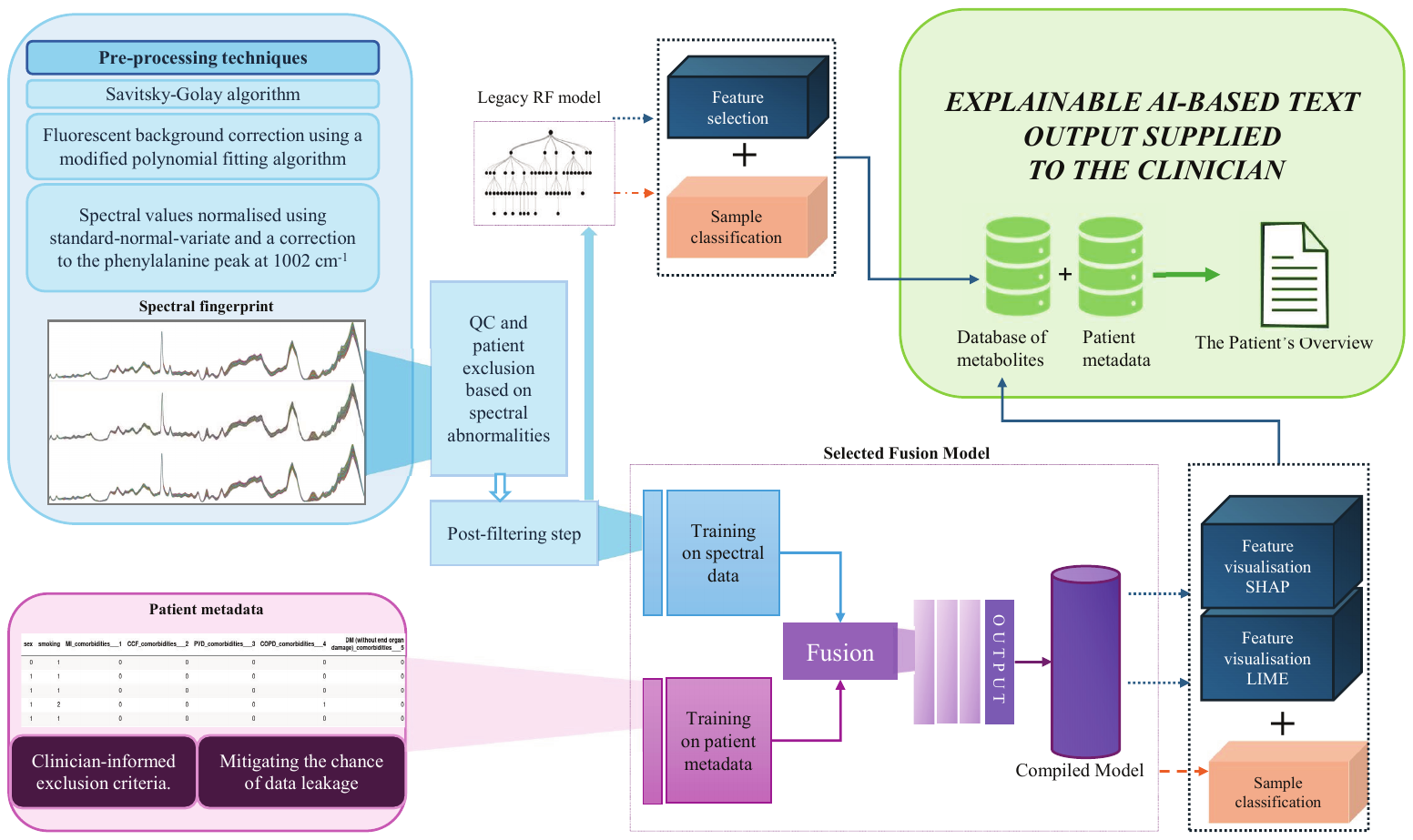} 
\caption{The outline of the proposed Clinical Expert system: Step 1 and 2 preprocessing the spectra and applying the exclusion criteria based on the metadata, Step 3 metadata of interest is supplied to Fusion Models, Step 3a Spectral values only are put through the RF model, Step 4 Important features are extracted from both, Step 3 and 3a, Step 5 Text output supplied to the clinician.}
\label{MODEL}
\end{figure*}

We divide our solution into five major tasks, as shown in Figure \ref{MODEL}:
\begin{itemize}
\item \textbf{Steps 1 and 2}: Independently preprocess and evaluate the patients' metadata and the spectral readings. Remove any spectral readings from the model if they are significantly different from the baseline spectral reading.
\item \textbf{Step 3}: Analysing the sample in the CRC and the Polyp Fusion Models.
\item \textbf{Step 3a}: Input the patient sample of interest into the legacy random forest (RF) model created using spectral values only.
\item \textbf{Step 4}: Explainability - Extract features of interest and sample classification from both the RF and Fusion Models.
\item \textbf{Step 5}: Convert the output into text that explains the results to the clinician. This text compares the spectral features between the patients, assigning them the putative chemical groups (e.g., sterols) for each model, and specifies which metadata were most relevant for this patient and how that compared to the rest of the samples in the model.
\end{itemize}

In Step 1, the spectral signal was subjected to standard Raman spectral preprocessing, which involved a) the Savitzky-Golay algorithm \( \hat{y}_n = \sum_{i=-m}^{m} c_i y_{n+i} \), where \( \hat{y}_n \) represents the smoothed spectral data at point \( n \), \( y_{n+i} \) are the original spectral data points within the window, and \( c_i \) are the filter coefficients, b) fluorescent background correction

\(
y_{\text{corr}} = y_{\text{meas}} - f(x) \quad \text{where} \quad f(x) = \sum_{i=0}^{n} a_i x^i
\); \( y_{\text{corr}} \) represents the corrected Raman intensity, \( y_{\text{meas}} \) the raw measured Raman intensity, \( f(x) \) denotes the polynomial fitting the fluorescent background model, c) standard cosmic ray removal procedures where
\[
y_{\text{corr}}(n) = 
\begin{cases}
y_{\text{meas}}(n), & \text{if } y_{\text{meas}}(n) < T \\
\frac{y_{\text{meas}}(n-1) + y_{\text{meas}}(n+1)}{2}, & \text{if } y_{\text{meas}}(n) \geq T
\end{cases}
\] 
where \( T \) is defined such that any measured intensity above this value is considered to be affected by cosmic rays, d) normalisation to the phenylalanine peak where the intensity at each wavenumber \( n \) is calculated using the equation \( I'_{n} = \frac{I_{n}}{I_{\text{Phe}}} \cdot C \), where \( I'_{n} \) is the normalized intensity, \( I_{n} \) is the original intensity, \( I_{\text{Phe}} \) is the intensity of the phenylalanine peak, and \( C \) is the scaling factor.

 We calculated the baseline for the patients, defined as the values within the mean for any condition, where the baseline for the condition was given by 
\(
B = \frac{1}{N} \sum_{n=1}^{N} y_n
\), 
where \( N \) is the total number of spectral measurements in the specific condition and \( y_n \) represents the intensity values of the preprocessed spectral data for this condition at each measurement point. Any sample that significantly diverged from that baseline was flagged as requiring greater attention from the clinician, since the result was drastically different from those of other patients, indicating a severe underlying problem. Any spectral readings that were significantly different from the baseline spectral reading were removed from the model. The number of samples that passed our pre- and post-filtering stages is listed in Tables \ref{tab:diagnosis_distribution} and \ref{tab:balan_diagnosis_distribution}.

In Step 2, the patient metadata collected by the clinician were investigated to exclude any potential data leakage in the model. This involved removing any metadata related to comorbidities that could explain why the patient is likely to suffer from polyps/CRC, e.g., Lynch syndrome \citep{Peltomki2023}. The metadata included a list of all diagnosed comorbidities, all medications taken, and other demographic information. In Step 3, the Fusion Model was trained on both the metadata and the spectral information. Step 3a involved inputting the sample of interest into a legacy RF model that had been trained solely on the spectral signal. This legacy model was split into training, validation, and testing (70\%, 10\%, 20\%). The model’s hyperparameters were fine-tuned to improve performance. The trained model was subjected to stratified k-fold cross-validation, as well as leave-one-out cross-validation (LOOCV), and cross-validation with an independent cohort of patients.

In Step 4, we compute Shapley Additive exPlanations (SHAP) values and LIME values, and extract the sample classification results for the Fusion Model. These results, along with the features of interest and the sample classification for the RF model, will be included in the report. Both the features of interest and the classification results will be presented in the report.

In Step 5, we provide a textual explanation of the model’s performance based on the additional feature annotations created with a specialist-in-the-loop.
\subsection{Data}
We have two data matrices; 1. Spectral Data: \( X_s \in \mathbb{R}^{n \times d_s} \), where \( n \) is the number of samples and \( d_s \) is the number of features in the spectral data; 2. Medical Records: \( X_m \in \mathbb{R}^{n \times d_m} \),  \( d_m \) is the number of features in the medical records. The data matrices are scaled and processed \(\tilde{X}_s = \operatorname{scale}(X_s)\) and \(\tilde{X}_m = \operatorname{scale}(X_m)\). 

\subsection{Fusion architectures}
\subsubsection{Early-fusion (feature level fusion).}
Early fusion involves combining feature vectors from various data modalities into a single vector, simplifying the process by requiring only the training of one model \citep{Steyaert2023}. The two scaled matrices are horizontally concatenated to form a single feature matrix,
\(X_{\text{c}} = \left[\tilde{X}_s \; \middle| \; \tilde{X}_m\right] \in \mathbb{R}^{n \times (d_s + d_m)}\). The input layer accepts \( X_{\text{c}} \): \(\mathbf{z}_0 = X_{\text{c}}\). 
For each hidden layer \( l \), the transformation is given by \(\mathbf{z}_{l+1} = \sigma(\mathbf{W}_l \mathbf{z}_l + \mathbf{b}_l)\), where \( \mathbf{W}_l \) is the weight matrix for layer \( l \), \( \mathbf{b}_l \) is the bias vector for layer \( l \), \( \sigma \) is the activation function, and \( \mathbf{z}_{l+1} \) is the output of the layer. 

\subsubsection{Joint fusion (intermediate fusion).} Intermediate fusion does not merge input data; instead, it utilises inference algoritms in order to generate a a joint multimodal low-level representation of the features, simultanously retaining the properties and the signal of individual modalities \cite{Steyaert2023}. In this early feature-level fusion
model, spectral and medical data are processed through separate dense layers, 
resulting in feature matrices \( \mathbf{F}_s \) and \( \mathbf{F}_m \), respectively. These extracted features are concatenated \( \mathbf{F}_c = \left[\mathbf{F}_s \; \middle| \; \mathbf{F}_m\right] \in \mathbb{R}^{n \times (d_s + d_m)} \). The combined features \( \mathbf{F}_c \) are then passed through additional dense layers with dropout for regularisation, where the transformation for each layer \( l \) is again given by \( \mathbf{z}_{l+1} = \sigma(\mathbf{W}_l \mathbf{z}_l + \mathbf{b}_l) \).
The final output layer generates a binary classification probability using a sigmoid function. 

\subsubsection{Late Fusion Model (Score Level Fusion).}
Late fusion operates by aggregating predictions at the decision level rather than combining features at the input stage \citep{Steyaert2023}. In this approach, individual models are trained on different data modalities, and their outputs are combined to make the final decision. The activation function is denoted by \( \sigma(\cdot) \), where ReLU is used for hidden layers and sigmoid for the output layer. The combined predictions from these models are then passed through additional dense layers, and the final output layer produces the probability for binary classification.

\subsubsection{Further stages.}
Fusion models were trained by minimizing the binary cross-entropy loss between the predicted probabilities \( \hat{\mathbf{y}} \) and the true labels 
\begin{equation}
\mathcal{L}(\mathbf{y}, \hat{\mathbf{y}}) = - \frac{1}{m} \sum_{i=1}^{m} \left[ y_i \log(\hat{y}_i) + (1 - y_i) \log(1 - \hat{y}_i) \right].
\label{eq:loss_function}
\end{equation}
The model's parameters for each layer are updated to minimize the loss function \( \mathcal{L} \) using the Adam optimizer, as shown in Equation 1. The model is validated on the validation set using accuracy. The ROC curve and AUC are computed to evaluate the model's performance.
The entire process can be summarized as follows,
\begin{align}
\notag
\text{Input: } X_{\text{combined}} &\rightarrow \text{NN: } \hat{y} = f(X_{\text{combined}}; \mathbf{W}, \mathbf{b}) \\
\text{Loss: } \mathcal{L}(y, \hat{y}) &\rightarrow \text{Minimize: } \mathcal{L} \notag
\end{align}
where \( f \) is the neural network function defined by the layers and activations. The goal is to find the optimal weights \( \mathbf{W} \) and biases \( \mathbf{b} \) that minimize the loss \( \mathcal{L} \).

\subsection{The summary of the legacy RF model}
\begin{table*}[t]
\caption{Evaluation metrics for control and disease groups in the legacy RF model}
\label{evaluation-metrics}
\centering
\begin{tabular}{lccccccc}
\hline
\textbf{Model} & \textbf{Label} & \textbf{Precision} & \textbf{Recall} & \textbf{F1} & \textbf{Accuracy} & \textbf{5-fold CV} & \textbf{LOOCV} \\
\textbf{} & \textbf{} & \textbf{} & \textbf{} & \textbf{} & \textbf{} & \textbf{Precision \(\mu\)} & \textbf{Precision} \\
\hline
RF Women & Control & 0.83 & 0.83 & 0.83 & \multirow{2}{*}{80.95\%} & 0.84 ± 0.046 & 88.6\% ± 9.48\% \\
& Disease & 0.78 & 0.78 & 0.78 &  &  &  \\
\hline
RF Men & Control & 0.72 & 0.95 & 0.82 & \multirow{2}{*}{82.00\%} & 0.83 ± 0.054 & 84.6\% ± 4.2\% \\
& Disease & 0.95 & 0.71 & 0.82 &  &  &  \\
\hline
RF Both & Control & 0.75 & 0.93 & 0.83 & \multirow{2}{*}{81.32\%} & 0.84 ± 0.04 & 90.04\% ± 6.63\% \\
& Disease & 0.92 & 0.70 & 0.80 &  &  &  \\
\hline
\end{tabular}
\end{table*}
The RF model was trained on spectra only. We created 3 different models (Table \ref{evaluation-metrics}). Both sexes: accuracy of 81.32\%, 0.87 ROC, and 92.00;
2) Men-only model: the accuracy of 82.00\%, ROC reached 0.90, and the precision was equal to 95.0\%. The stratified
k-fold cross-validation mean ROC score in men was equal
to 0.83 $\pm$ 0.054, with a mean accuracy of 76.6;
3) Women only: the accuracy was 80.95\%, ROC 0.88, and
precision 78.0\%. A k-fold cross validation resulted in a
mean ROC 0.84 $\pm$ 0.046, a mean accuracy equal to 80.4%
$\pm$ 5.04\% and a mean precision 88.6\% $\pm$ 9.48\%. For
LOOCV, the mean ROC score was 0.782, with mean accuracy equal to 80.1\%, and precision 87.7\%.

\subsection{SHAP and LIME}
SHAP (SHapley Additive exPlanations), is a game theoretic-based approach derived form Shapley values \cite{shapley:book1952}, which explains the predictions derived from the model by treating each feature as a player and the model's output as the game's payoff \cite{Kariyappa2024}.  SHAP assigns a score to every feature fed into the model by measuring a marginal contribution of features that constitute different coalitions, estimating the feature’s contribution to the model’s prediction \cite{Kelodjou2024}.  

Assuming a set of features \(N = \{1, 2, \ldots, M\}\), the SHAP value 
\(\phi_i\)  for the \({i}\)-th feature belonging to the input \(x\) in the model \(f\) is derived by computing the weighted average of the change in in \(f\)'s predictions when \({i}\) is added to a subset of features \(S\) as outlined in Equation \ref{eq:shap}.

\begin{equation}
\phi_i(x, f) = 
\begin{aligned}[t]
&\sum_{S \subseteq N \setminus \{i\}} \frac{|S|!(M - |S| - 1)!}{M!} \\
&\quad \times \left[ f(x_{S \cup \{i\}}) - f(x_S) \right].
\end{aligned}
\label{eq:shap}
\end{equation}

Although methods such as Kernel SHAP are widely used, the relevance of the explanations is often diminished due to Kernel SHAP's instability - different executions of a model can result in inconsistent explanations \cite{Kelodjou2024}
Kernel SHAP's instability arises from its stochastic neighbour selection procedure. 

The practical implementation of Kernel SHAP estimates approximate scores via a linear regression from perturbed samples to manage the computational cost. Perturbation-based methods often suffer from instability issues, affecting reproducibility. 

Although local post-hoc perturbation based-methods (such as LIME and SHAP) are widely used to explain black box models, literature
highlights stability/reproducibility as a critical property, proposing a framework to determine sufficient perturbation points for stable explanations \cite{Kelodjou2024, https://doi.org/10.48550/arxiv.2106.07875}. 
\subsubsection{LIME} - Local Interpretable Model-Agnostic Explanations, approximates the behavior of a model \(f\) locally around a specific instance \(x\) with a simpler, interpretable model \(\hat{f}\) 
\cite{https://doi.org/10.48550/arxiv.1602.04938}. The objective of LIME can be formulated as shown in Equation 3:
\begin{equation}
   \hat{f}(x) = \arg\min_{g \in G} \sum_{x' \in \mathcal{D}} \pi_x(x') \left( f(x') - g(x') \right)^2 + \Omega(g).
\label{eq:LIME} 
\end{equation}

where \(G\) represents the family of interpretable models. The set \(\mathcal{D}\) is generated by perturbing the original instance \(x\), and \(\pi_x(x')\) is a proximity measure that assigns higher weights to perturbed instances \(x'\) closer to \(x\). The term \(\Omega(g)\) penalizes the complexity of the interpretable model \(g\), encouraging simpler explanations. LIME produces an interpretable surrogate model \(g\) capable of approximating the local behaviour of the original model, estimating the contribution of features to the final prediction\cite{NEURIPS2023_71ed0429}. Literature reports LIME having poor local fidelity and instability \cite{NEURIPS2023_71ed0429}.     
To address the issues with perturbation-based xAI methods, we identified features unique for each class fed into the model, which were deemed important by both, LIME and SHAP, and we provided the clinician with a summary of the results.

\subsection{Human in the Loop}
To mitigate model biases and incorporate expert opinion, we created a library of annotations for spectral features corresponding to specific Raman shifts, considering molecular functional groups as distinct units. This provides information on compounds with higher or lower presence in the sample. We excluded any comorbidities where patients are at a higher risk of developing polyps/CRC, and clinicians excluded patients from sensitive groups, such as those struggling with substance abuse. Prescription medications were considered acceptable since the dosage is known.

Additionally, we created a library of metabolites altered in comorbidities listed in the patient metadata, taking into consideration metabolic pathways altered. Furthermore, we created a literature-based library, where we listed metabolites linked to the presence of polyps or colorectal cancer. 

\subsection{Text Output}
In the final stage, we produce a text summary that briefs the clinician on all model findings (Appendix A), and summarises the library annotations and metadata information based on SHAP and LIME values. Thanks to the libraries summarising metabolites changes in comorbidities and polyps/CRC, the clinical Expert system summarises the overlap between the two. The clinician receives information on patient classification in each model.

\section{Experiments and Results}
We conducted several experiment setups: models built on the unbalanced dataset, the balanced dataset, a late-fusion model, a joint fusion model, and an early fusion model. \subsection{Polyp-Only Model Performance}

The results from the polyp-only model show notable differences in performance across the fusion techniques (Table 4).

The \textbf{Early Fusion model} performs significantly better than the baseline \textbf{Vanilla ANN Keras model}, achieving an accuracy of 0.878, a precision of 0.731, and a recall of 0.594, leading to an F1 score of 0.655. The AUC for this model is 0.861 (Figure 2), indicating a strong ability to differentiate between classes.

The \textbf{Joint Fusion model} outperforms the others, with an accuracy of 0.896, a precision of 0.815, and a recall of 0.688. It achieves the highest AUC of 0.887 and an F1 score of 0.746, making it the best-performing model in the polyp-only detection task.

\begin{figure*}[t]
\centering
\includegraphics[width=\textwidth]{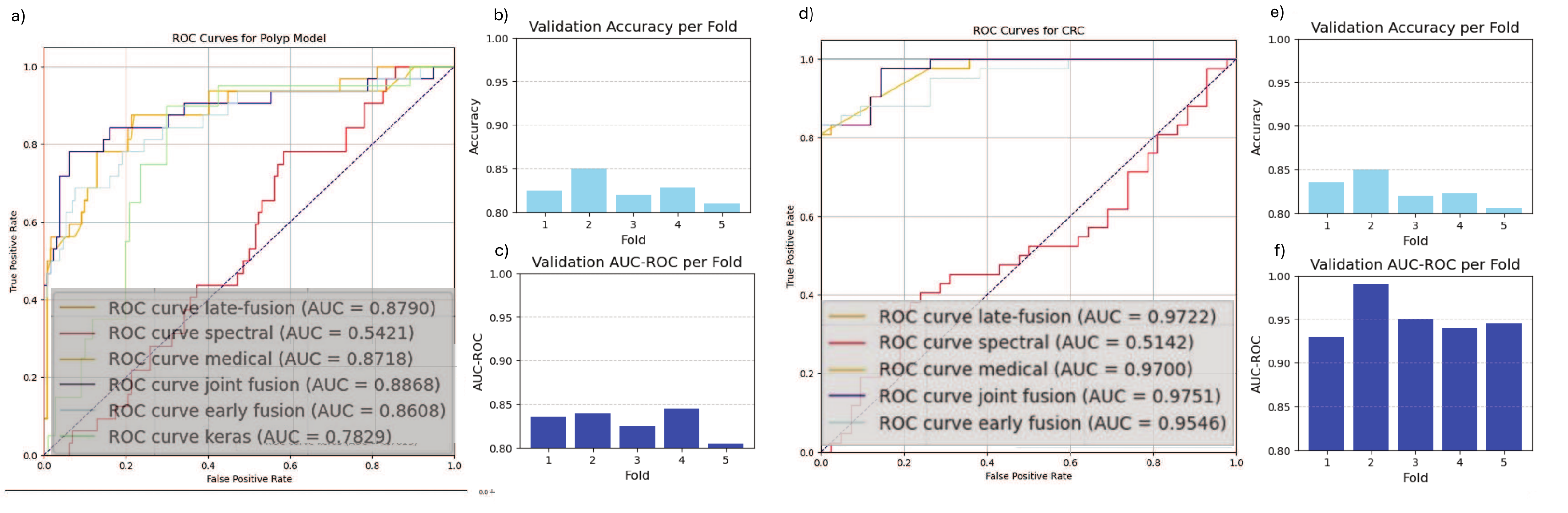}
\caption{Performance metrics in the model trained on dataset containing controls and polyps: a) The AUC performance on the balanced out model in polyp-control dataset, b) Validation accuracy per Fold, c) Validation AUC-ROC per fold. Performance metrics in the model trained on dataset containing controls and CRCs: d) The AUC performance on the balanced out model in polyp-control dataset, e) Validation accuracy per Fold, f) Validation AUC-ROC per fold}
\label{fig1}
\end{figure*}

\begin{figure*}[t]
\centering
\includegraphics[width=\textwidth]{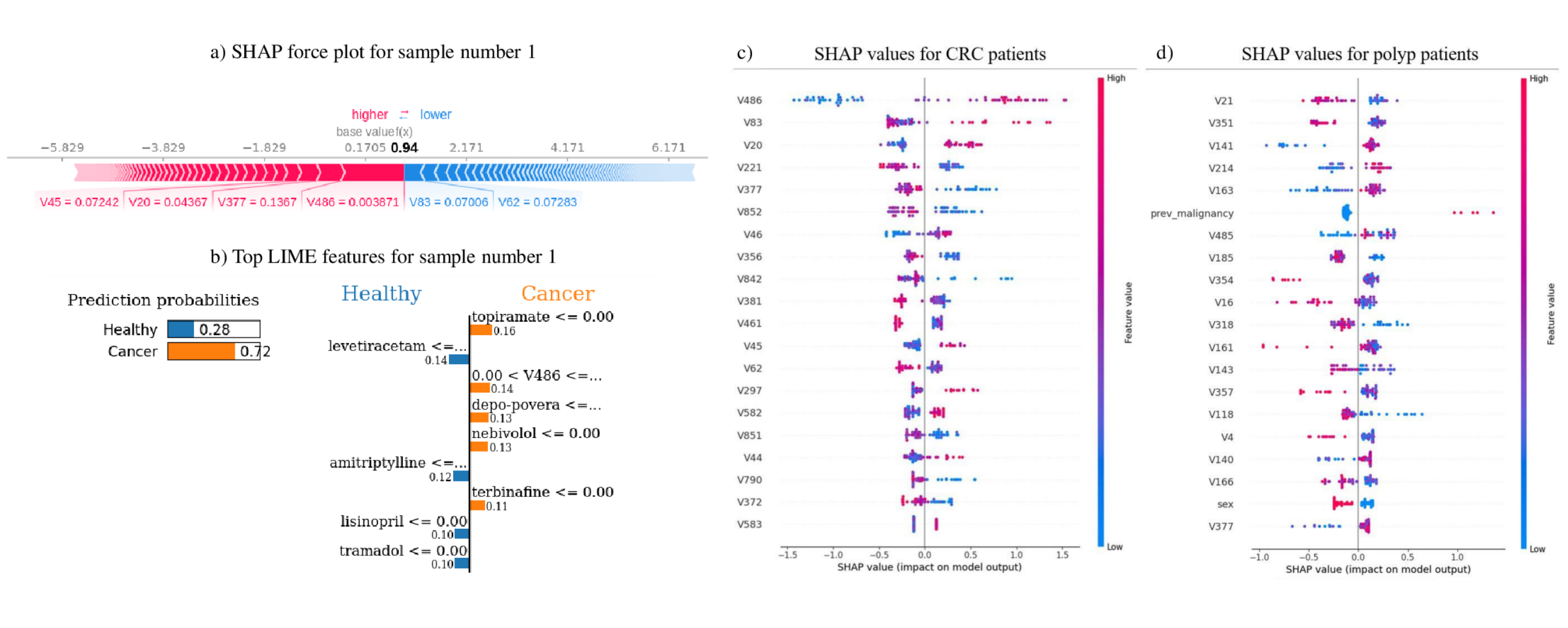} 
\caption{SHAP and LIME values across the top performing models. Panel a) and b) present SHAP and LIME feature importance for a patient suffering from CRC. Panel c) the top SHAP values for the model trained on controls and CRC patients, d) the top SHAP values for the model trained on controls and polyps}
\label{shap}
\end{figure*}

The \textbf{Late Fusion model} shows a high precision of 0.833 but suffers from a low recall of 0.156, resulting in a poor F1 score of 0.263 despite a high AUC of 0.879. This suggests that while it is precise in its predictions, it fails to identify a significant portion of polyp cases.

\begin{table}[t]
\fontsize{9}{11}\selectfont
\setlength{\tabcolsep}{1mm}  
\caption{Model performance for the polyp-only model}
\label{evaluation-metrics-polyp}
\centering
\begin{tabular}{lccccc}
\hline
\textbf{Model} & \textbf{Accuracy} & \textbf{Precision} & \textbf{Recall} & \textbf{AUC} & \textbf{F1 Score} \\
\hline
Vanilla Keras         & 0.732  & 0.25   & 0.05   & 0.782  & 0.083  \\
Early Fusion  & 0.878  & 0.731  & 0.594  & 0.861  & 0.655  \\
Joint Fusion  & 0.896  & 0.815  & 0.688  & 0.887  & 0.746  \\
Late Fusion   & 0.829  & 0.833  & 0.156  & 0.879  & 0.263  \\
\hline
\end{tabular}
\end{table}

\subsection{CRC-Trained Model Performance}

The CRC-trained model demonstrates stronger performance across all fusion methods (Table \ref{evaluation-metrics-crc}) compared to the polyp-only model. The \textbf{Early Fusion model} achieves the highest accuracy of 0.976, with precision, recall, and F1 score all at 0.881. Its AUC is slightly lower at 0.955 compared to other methods but still indicates strong classification capability.

The \textbf{Joint Fusion model} follows closely with an accuracy of 0.972 and matches the Early Fusion model in precision, recall, and F1 score (all at 0.881). However, it boasts the highest AUC of 0.975, indicating slightly better discrimination between cancerous and non-cancerous cases.

The \textbf{Late Fusion model} performs comparably in terms of precision (0.884) and recall (0.905), with an F1 score of 0.894. Its accuracy is slightly lower at 0.893, but the AUC remains high at 0.972 (Figure \ref{fig1}d), suggesting that while it has slightly lower accuracy, it is still effective at distinguishing CRC cases.

\begin{table}[t]
\fontsize{9}{11}\selectfont
\setlength{\tabcolsep}{1mm}  
\caption{Metric performance for the CRC-trained model}
\label{evaluation-metrics-crc}
\centering
\begin{tabular}{lccccc}
\hline
\textbf{Model} & \textbf{Accuracy} & \textbf{Precision} & \textbf{Recall} & \textbf{AUC} & \textbf{F1 Score} \\
\hline
Joint Fusion  & 0.972  & 0.881  & 0.881  & 0.975  & 0.881  \\
Early Fusion  & 0.976  & 0.881  & 0.881  & 0.955  & 0.881  \\
Late Fusion   & 0.893  & 0.884  & 0.905  & 0.972  & 0.894  \\
\hline
\end{tabular}
\end{table}

The performance is presented in Figure \ref{fig1}. An example of SHAP values produced by the polyp-only model is present in Figure \ref{shap}d.

\subsection{Features of importance in SHAP and LIME}
In Figure \ref{shap}a and b, we present representative feature importance explanations for one patient using the best-performing model for each dataset. Figures \ref{shap}a and \ref{shap}b highlight that both LIME and SHAP identified feature V486 as significant in a patient diagnosed with colorectal cancer (CRC).
In Figure \ref{shap}c, the SHAP explanation indicates spectral values as the most important features, even when fusion techniques are employed. Conversely, in Figure \ref{shap}d, the top features include both, features form spectral data and patient metadata, emphasising previous malignancy and sex. This distinction may reflect the metabolic reprogramming characteristic of cancer \cite{Hanahan2022}, which manifests as significant alterations in spectral patterns.

For patients with polyps, sex and previous malignancy are likely selected as features contributory to model's decision-making, potentially indicating sex-specific hormonal influences or, in the case of prior malignancies, lifestyle factors, or inherited predispositions leading to mutations or cancer susceptibility \cite{Brennan2021}.

\section{Discussion}
Our framework outperformed current available blood-based CRC diagnostic tests. Tests such as Guardant Shield, the cfDNA blood-based test achieved a sensitivity of 83.1\% for colorectal cancer detection and 13.2\% for advanced precancerous lesions, with a specificity of 89.6\% for advanced neoplasia (colorectal cancer or advanced precancerous lesions) \cite{Chung2024}. In comparison to our proposed framework, the cfDNA test’s low sensitivity for advanced precancerous lesions limits its utility for early-stage interventions \cite{Chung2024}. Our method addresses these limitations by offering enhanced performance for detecting advanced precancerous lesions while maintaining good metrics performance, providing a more robust solution for early CRC screening in average-risk populations.

Although the framework proposed here does not outperform colonoscopy - the gold standard for CRC diagnostics \cite{Shaukat2022}, it could be utilised as a screening test in the general population. 
The need for cost-effective and rapid screening tests, with results that can be easily interpreted by clinicians, is particularly urgent given the rise in CRC among younger patients who would not typically undergo routine colonoscopy screening \cite{Constantinou2023}.

Utilising a clinical expert system that integrates multimodal patient data with Raman spectroscopy enables a noninvasive framework that is less time-consuming for clinicians, requires less highly qualified medical staff, improves patient compliance, and bridges the gap between the model’s outputs and medical staff.

\section{Conclusion} 
While our approach offers promising advancements in the early detection of CRC and polyps, several limitations must be acknowledged:
\begin{itemize}
\item \textbf{Dataset Bias and Generalisability:} Our dataset may contain inherent biases due to the specific demographic and geographic characteristics of the patient population. These biases could affect the model’s generalisability to broader, more diverse populations. Additionally, the relatively small sample size may limit the robustness of our findings. However, the pre-and post-processing applied in this model provides a certain level of safeguarding. If the spectra fed into the model is drastically different from our baseline, it will be rejected and marked as unsuitable for our frameworks. 
\item \textbf{Comorbidity Exclusion:} By excluding patients with certain comorbidities and those from sensitive groups, the model’s applicability may be restricted. 
\item \textbf{Spectral Data Variability:} Although we ensured that the quality and consistency of Raman spectral in our model was optimised, it is plausible that data produced in a different laboratory could not meet our pre- and post-processing quality control (QC). 
\item \textbf{Model Interpretability:} The text format provides the clinician with an explanation of the model; however, due to the nature of Raman spectra readings, it primarily includes a summary of potential functional groups present in the sample. We provide a list of plausible chemicals found in the sample, but this requires the clinician to have some level of biochemical knowledge.
\item \textbf{Limited Clinical Validation:} Our group has validated the model performance in the clinical setting within the local NHS practices, however, a larger and more varied patient population could confirm the model's efficacy and reliability.
\end{itemize}

This study introduces a novel method for early detection of colorectal cancer (CRC) and polyps using Raman spectroscopy combined with machine learning and explainable AI. Our approach integrates spectral data with patient metadata, enhancing diagnostic accuracy and providing clear, actionable insights for clinicians.

While the model shows promise, limitations such as dataset biases and the need for broader clinical validation must be addressed. Ethical considerations, including patient privacy and equitable access, are crucial.

Future work should focus on validating the model in diverse settings and expanding the dataset. This research provides a foundation for advancing cancer diagnostics through the integration of AI and spectroscopy, aiming for more effective and personalised detection methods.
\section*{Ethical Statement}
This work has the potential to make early colorectal cancer detection available to a greater number of the population through an acceptable blood test as an alternative to current faecal methods, which have lower uptake and compliance from underserved communities. The ability to accurately detect pre-malignant features from a blood test has the potential to make screening programmes more effective and offers the possibility of colorectal cancer prevention.

The authors declare the following financial interests/personal relationships which may be considered as potential competing interests: PRD and DAH declare their involvement in CanSense Ltd (company number 11367637), CanSense Group Ltd (12346893) and CanSense Laboratories Ltd (15380816). No other authors have any competing interests to declare.

\section*{Acknowledgments}
This work is supported by the UKRI AIMLAC CDT, funded by grant EP S023992 1; Cancer Research Wales; and Health and Care Research Wales.
We would like to thank Swansea Bay University Health Board as a sponsor of the studies and the study participants who gave their permission for their blood samples to be used.
Furthermore, we would like to thank our funding bodies and all members of our laboratory who made this work possible through their contributions.
We would like to acknowledge Prof. Gert Aarts for his invaluable support. 
\section*{Author Contributions} SY supervised the project. NS conducted ML experiments, created figures, analyzed data, produced the outputs, and drafted the manuscript. NS and SY developed the study design. NS, PD, and SY interpreted the results. DH and PD secured funding. NS collected the data, with the help of AMT. NS and RLM conducted the literature review, and finalized the manuscript. RLM assisted in figure creation and write-up of manuscript sections. AMT recruited the patients. DAH and AMT provided clinical advice.

\bibliography{CameraReady/LaTeX/aaai25}

\appendix
\section{TECHNICAL APPENDIX}\label{sec:appendix}
\subsection{A. CLINICAL SYSTEM TEXT OUTPUT}\label{sec:clinical_output}
The patient reported to the clinic is XX years old, BMI XX, which is 2\% higher than the recommended weight,
with a positive smoker status, identifies as a male. The patient's NHS number is XXXXXXXXXX.

The patient suffers from hypertension, and had not previous reported malignancy.

Before the test, the patient received bowel prep. Medical metadata reports no diagnosis of asthma, hypothyroidism, hyperthyroidism comorbidities, atrial fibrillation comorbidities, IHD, 
anxiety and/or depression, hypercholesterolaemia, arthritis. No additional symptoms were reported, including no gastrointestinal bleeding,  weight loss, loss of appetite, change in bowel habit, abdominal pain, abdominal mass,
anal pain, anal lump/mass, rectal mass, new anaemia, looser stool, change in bowel habit, increased frequency in bowel habit, urgency change, incomplete emptying, constipation. 
Patient history excluded comorbidities include: diverticular disease, haemorrhoids, inflammatory bowel disease, microscopic colitis, proctitis, angiodysplasia, hyperplastic polyps. The patient reported taking the following
medications recently: bendroflumathiazide, hypromellose 0.3\% eye drops, paracetamol.  The Polyp model recommended that the patient should have a colonoscopy, the CRC model did not recommend the patient for colonoscopy. Therefore, 
based on the Raman spectra and the patient metadata, the patient is medium risk for developing/suffering from CRC. The SHAP values for metadata flagged up 1 value out of 6 positives in the metadata (among 701 possible 
features that patients were investigated for). SHAP values for the spectral dataset flagged up 6 features (V160, V333, V312, V577, V46, V21). Patients suffereing from hypertension tend to report changes in the metabolism of 
monosaccharides and disaccharides, such as galactose, glucosamine, and sucrose, Amino acids: Increased levels of aminobenzoic acid, daminozide, organic acids: Increased levels of aminoacids, hydroxyacids, and ketoacids, 
steroids and fatty acyls, Gut microbial metabolites: Acetate and butyrate, 13-HODE and 9-HODE, DMTPA: amino acid metabolite associated with renal function. 
The results showed activation in the pyruvate metabolism and glycerolipid metabolism, suggesting a presence of the polyp. Peaks suggesting aacetate and glycerol were altered, suggesting changes related to he presence of polyps. 
No peaks related to changes in the glycolysis and glycine, serine, and threonine metabolism or lactate and citrate were shown, suggesting no presence of CRC and alterations of TCA cycle.

Overall, the model flagged up the following features:
peaks suggesting the presence of a polyp: 6/10 (60.0
peaks suggesting the presence of CRC: 0/12 (0.0
features which could be false positives due to medications/comorbidities: 2
names of features leading to potential false positives: acetate, organic acids

\section*{Patient Information}
\begin{itemize}
    \item Age: XX years
    \item BMI: XX (2\% higher than the recommended weight)
    \item Gender: Male
    \item Smoker Status: Positive
    \item NHS Number: XXXXXXXXXX
\end{itemize}

\section*{Medical History}
\begin{itemize}
    \item Current conditions: Hypertension
    \item No history of malignancy
    \item No diagnoses of the following:
    \begin{itemize}
        \item Asthma
        \item Hypothyroidism
        \item Hyperthyroidism
        \item Atrial fibrillation
        \item Ischemic heart disease (IHD)
        \item Anxiety or depression
        \item Hypercholesterolemia
        \item Arthritis
    \end{itemize}
    \item No additional symptoms reported:
    \begin{itemize}
        \item Gastrointestinal bleeding
        \item Weight loss
        \item Loss of appetite
        \item Changes in bowel habits
        \item Abdominal pain or mass
        \item Anal pain or lump/mass
        \item Rectal mass
        \item New anemia
        \item Loose stools
        \item Increased frequency or urgency of bowel movements
        \item Incomplete emptying
        \item Constipation
    \end{itemize}
    \item Excluded comorbidities:
    \begin{itemize}
        \item Diverticular disease
        \item Hemorrhoids
        \item Inflammatory bowel disease
        \item Microscopic colitis
        \item Proctitis
        \item Angiodysplasia
        \item Hyperplastic polyps
    \end{itemize}
\end{itemize}

\section*{Medications}
\begin{itemize}
    \item Bendroflumethiazide
    \item Hypromellose 0.3\% eye drops
    \item Paracetamol
\end{itemize}

\section*{Risk Assessment}
\begin{itemize}
    \item Polyp Risk Model: Recommended colonoscopy
    \item CRC Risk Model: Did not recommend colonoscopy
    \item Classification: Medium risk for CRC based on Raman spectra and metadata
\end{itemize}

\section*{SHAP Analysis}
\begin{itemize}
    \item Metadata: 1 significant feature flagged out of 6 positives (701 features evaluated)
    \item Spectral Dataset: 6 significant features flagged: V160, V333, V312, V577, V46, V21
\end{itemize}

\section*{Metabolic Observations}
Patients with hypertension tend to exhibit the following metabolic changes:
\begin{itemize}
    \item \textbf{Monosaccharides and disaccharides:} Elevated galactose, glucosamine, and sucrose
    \item \textbf{Amino acids:} Elevated aminobenzoic acid and daminozide
    \item \textbf{Organic acids:} Elevated amino acids, hydroxyacids, and ketoacids
    \item \textbf{Steroids and fatty acyls}
    \item \textbf{Gut microbial metabolites:} Elevated acetate, butyrate, 13-HODE, and 9-HODE
    \item \textbf{DMTPA:} Associated with renal function
\end{itemize}

\section*{Test Results}
\begin{itemize}
    \item Activation of pyruvate and glycerolipid metabolism suggests the presence of a polyp
    \item Altered peaks associated with acetate and glycerol indicate changes related to polyps
    \item No observed peaks associated with:
    \begin{itemize}
        \item Glycolysis
        \item Glycine, serine, and threonine metabolism
        \item Lactate and citrate
    \end{itemize}
    \item No evidence of CRC or alterations in the TCA cycle
\end{itemize}

\section*{Summary of Results}
\begin{itemize}
    \item Peaks suggesting the presence of a polyp: 6/10 (60.0\%)
    \item Peaks suggesting the presence of CRC: 0/12 (0.0\%)
    \item Potential false positives due to medications/comorbidities: 2
    \item Features potentially leading to false positives: Acetate and organic acids
\end{itemize}

\end{document}